\documentclass{article}

\usepackage{microtype}
\usepackage{graphicx}
\usepackage{subfigure}
\usepackage{booktabs} %

\PassOptionsToPackage{hyphens}{url}
\usepackage{hyperref}

\usepackage[accepted]{icml2023_deployableGenAI}
\usepackage{amsmath}
\usepackage{amssymb}
\usepackage{mathtools}
\usepackage{amsthm}

\usepackage[capitalize,noabbrev]{cleveref}

\theoremstyle{plain}

\theoremstyle{definition}

\theoremstyle{remark}

\usepackage[textsize=tiny]{todonotes}

\usepackage{paralist}
\usepackage[belowskip=1pt,aboveskip=0pt]{caption}

\newcommand{\descr}[1]{\noindent\textbf{#1}}

\icmltitlerunning{Privacy-Preserving Synthetic Data in the Enterprise}

\begin{document}

\twocolumn[
\icmltitle{On the Challenges of Deploying\\Privacy-Preserving Synthetic Data in the Enterprise}

\icmlsetsymbol{equal}{*}

\begin{icmlauthorlist}
\icmlauthor{Lauren Arthur}{hazy}
\icmlauthor{Jason Costello}{hazy}
\icmlauthor{Jonathan Hardy}{hazy}
\icmlauthor{Will O’Brien}{hazy}
\icmlauthor{James Rea}{hazy}
\icmlauthor{Gareth Rees}{hazy}
\icmlauthor{Georgi Ganev}{hazy,ucl}
\end{icmlauthorlist}

\icmlaffiliation{hazy}{Hazy, London, UK}
\icmlaffiliation{ucl}{University College London, London, UK}

\icmlcorrespondingauthor{Georgi Ganev}{georgi@hazy.com}

\icmlkeywords{Machine Learning, ICML}

\vskip 0.3in
]

\printAffiliationsAndNotice{}  %

\begin{abstract}
Generative AI technologies are gaining unprecedented popularity, causing a mix of excitement and apprehension through their remarkable capabilities.
In this paper, we study the challenges associated with deploying synthetic data, a subfield of Generative AI.
Our focus centers on enterprise deployment, with an emphasis on privacy concerns caused by the vast amount of personal and highly sensitive data.
We identify 40+ challenges and systematize them into five main groups -- i)~generation, ii)~infrastructure \& architecture, iii)~governance, iv)~compliance \& regulation, and v)~adoption.
Additionally, we discuss a strategic and systematic approach that enterprises can employ to effectively address the challenges and achieve their goals by establishing trust in the implemented solutions.
\end{abstract}

\section{Introduction}
\label{sec:intro}

Recently, Generative AI has made significant advancements, with applications and capabilities spanning text, code, image, video, speech, and structured data~\cite{sequoia2022generative, gartner2023beyond}.
The acceptance of Generative AI products has reached unprecedented levels.
For instance, OpenAI's ChatGPT attracted an estimated 100M active users in January alone~\cite{reuters2023chatgpt}.
This wide-spread adoption, however, has not spread to large organizations as they feel unprepared~\cite{kpmg2023kpmg} while facing a plethora of concerns, including exposing themselves to ethical, security, privacy, robustness, copyright, compliance and legal risks~\cite{carlini2021extracting, weidinger2021ethical, cnn2023dont, entrepreneur2023history, techcrunch2023the}.
On the contrary, many companies have banned internal use of products like ChatGPT and GitHub Copilot~\cite{techcrunch2023apple}.

\descr{Motivation.}
As it is still unclear how/if said concerns can be resolved, we focus on synthetic data, a more established subfield of Generative AI.
On the one hand, synthetic data is produced by similar generative models, e.g., GANs~\cite{goodfellow2014generative}, Transformers~\cite{sohl2015deep}, and Diffusion Models~\cite{vaswani2017attention}.
On the other, it is typically trained on smaller-scale tabular datasets (vs. text/image), owned by a single entity (vs. public).
This ownership provides enhanced control over the entire data generation process, contributing to increased trustworthiness.
Moreover, synthetic data has attracted the attention of reputable organizations~\cite{rs2023privacy, un2023guide, oecd2023emerging} and regulators~\cite{ico2022privacy, fca2023synthetic}.
However, these studies lack in-depth insights into the practical challenges and considerations that companies encounter when deploying synthetic data solutions.

In the digital age, enterprises have accumulated and safeguarded immense volumes of their users' sensitive and personal data, making it one of most highly valued asset in their possession~\cite{nyt2018as, nyt2019twelve}.
Unfortunately, many organizations impede data access, maintain data silos, and discourage data sharing, which undermines efforts to capitalize on the potential business and social benefits~\cite{gartner2021data}.
Privacy-preserving synthetic data presents a promising solution that holds immense potential in addressing these challenges in an ethical and trustworthy way.

\begin{figure}
    \centering
    \includegraphics[width=0.95\linewidth]{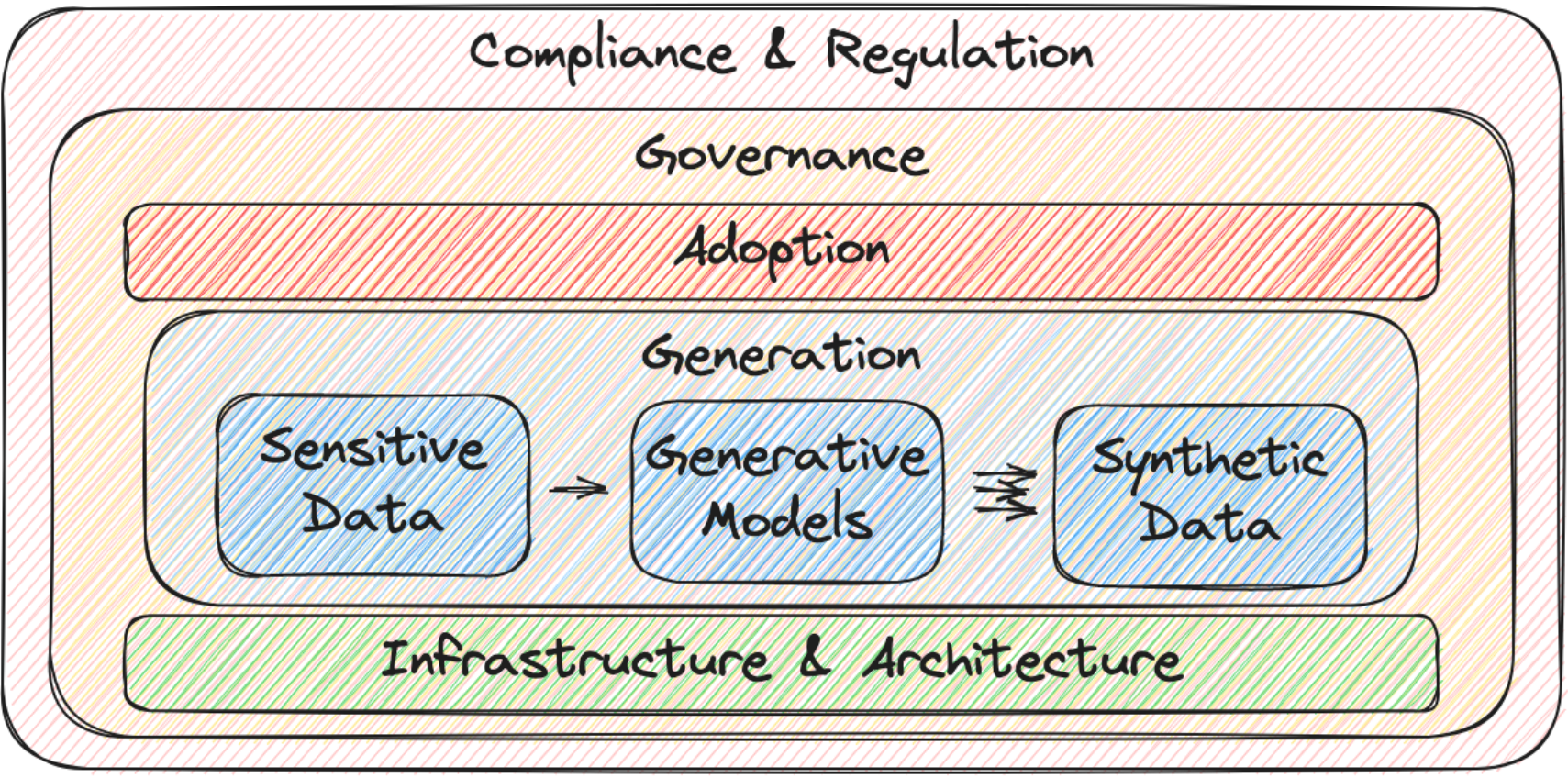}
    \caption{Main challenges of synthetic data deployment.}
    \label{fig:challenges}
    \vspace{-0.3cm}
\end{figure}

\descr{Main Contributions.}
Our findings can be summarized:
\begin{compactenum}
    \item We identify and explore 40+ challenges of deploying privacy-preserving synthetic data in large enterprises spanning beyond machine learning, unlike previous studies~\cite{assefa2020generating, jordon2022synthetic}.
    \item We categorize the challenges into five groups: generation, infrastructure \& architecture, governance, compliance \& regulation, and adoption (displayed in Fig.~\ref{fig:challenges}).
    \item We discuss the necessity of a structured approach that can help organizations frame the challenges and scale synthetic data deployment while establishing trust.
\end{compactenum}

In Sec.~\ref{sec:chall}, we discuss the main challenges and provide discussion in Sec.~\ref{sec:dis}, while in App.~\ref{sec:prelim}, we define some key concepts including synthetic data and Differential Privacy~(DP).

\section{Main Challenges}
\label{sec:chall}

In this section, we discuss the challenges associated with deploying synthetic data in an enterprise context.
We provide an overview of common considerations/obstacles and highlight those that are specifically relevant to enterprises.

\subsection{Generation}
\label{subsec:gen}
We examine the challenges of training, generation, evaluation from a technical perspective.
We discuss data/models operations and governance in Sec.~\ref{sec:inf} and~\ref{sec:gov}.

\descr{Data Preparation.}
Assuming there is data access, the following challenges need to be taken into consideration:
\begin{compactitem}
    \item {\em Data quality}: ensuring high-quality and diverse train data includes collecting rich up-to-date data from various sources, validating/cleaning it, maintaining consistent pipelines, integrating legacy formats/systems.
    \item {\em Data preprocessing}: essential preprocessing steps include standardizing and normalizing formats, addressing missing values and outliers, handling custom entities and business rules, labeling and annotating data, and conducting feature engineering across databases.
\end{compactitem}

Non-technology companies were lacking adequate investment in both areas just a few years ago~\cite{mckinsey2019driving}.

\descr{Generating Process.}
Even though synthetic data could be presented as a panacea solution, in fact, there is no one-model-fits-all use cases~\cite{tao2022benchmarking} as different generative approaches/models are better suited for different use tasks and settings~\cite{ganev2023understanding}.
Important challenges in selecting the best candidate model(s) include:
\begin{compactitem}
    \item {\em Data domain}: each data domain (e.g., single table, time-series/sequential, multi-table) presents its own unique challenges.
    While a wide range of solutions are available for the former~\cite{zhang2017privbayes, xie2018differentially, jordon2018pate, mckenna2021winning, mckenna2022aim}, there are only a few privacy-preserving options for the latter two~\cite{lin2020using, xu2023synthetic}.
    \item {\em Use case}: the use case with its associated task (e.g., capturing statistics, preserving query answers, classification, etc.) and complexity as well as the chosen evaluation criteria could limit the choice.
    \item {\em Domain expertise}: different domains might require (enterprise) specific knowledge to be encoded into the model to achieve higher utility/scalability.
    \item {\em Robustness}: edge cases, outliers require extra attention.
    \item {\em Models comparison}: due to the uncertainty/variability of how a model would perform a priori, often it is necessary to train and compare different models.
\end{compactitem}

\descr{Privacy.}
While applying DP provides formal mathematical privacy protections, it comes with its own challenges:
\begin{compactitem}
    \item {\em Threat model}: it is necessary to assess whether DP addresses the relevant threats -- it successfully defends vs membership/reconstruction inference attacks~\cite{hilprecht2019monte}.
    However, DP does not protect company-specific values such as categorical ids/names.
    \item {\em Privacy unit/budget}: consistent definition of the privacy unit is required across the entire end-to-end pipeline.
    Setting the privacy budget is context-specific and challenging~\cite{hsu2014differential}.
    Also, various DP mechanisms allocate their budget differently.
    \item {\em Utility effect}: DP usually reduces utility (unfortunately, disproportionally affecting outliers and minorities~\cite{stadler2022synthetic, ganev2022robin}).
    \item {\em Implementation}: Coding DP mechanisms is complex; practitioners should rely on proven/public solutions.
\end{compactitem}

\descr{Evaluation.}
Evaluation is also pivotal yet challenging area:
\begin{compactitem}
    \item {\em Quality}: trust requires measurability of the desired properties of synthetic data -- utility, fidelity, diversity, authenticity, fairness, etc.
    Not only are they hard to measure, but defining them is also challenging~\cite{alaa2022faithful, breugel2023beyond}.
    \item {\em Privacy and auditability}: auditing the model ensures the privacy of synthetic data by avoiding leakage of undesirable properties.~\cite{houssiau2022framework, houssiau2022tapas}.
\end{compactitem}

\citet{jordon2022synthetic, cummings2023challenges} provide further discussion on the technical challenges of DP synthetic data.

\subsection{Infrastructure \& Architecture}
\label{sec:inf}

Building on an enterprise platform entails requirements in system design and architecture, such as security, scalability, and distributed state management~\cite{fowler2012}.

\descr{Network Topologies.}
Deploying software in enterprise networks is challenging for various reasons:
\begin{compactitem}
    \item {\em Complexity}: unique organizational needs and structure drive networked system topology~\cite{bano2016empirical}.
    \item {\em Legacy}: achieving agility/change~\cite{leffingwell2007scaling} in legacy systems is challenging for new technologies.
    \item {\em Cloud multiplicity}: ``cloud transformation''~\cite{jamshidi2013, avram2014} brought various options to consider: on-premises, hybrid, public, and multi-cloud.
\end{compactitem}

\descr{Distributed Data.}
Data distribution across multiple at-rest locations poses significant barriers:
\begin{compactitem}
    \item {\em Segmentation}: organizational structure and data-centric regulations~\cite{pci2013} result in data spanning multiple security tiers and availability zones.
    \item {\em Discovery}: visibility of existent datasets for synthetic data use can become non-trivial at enterprise scale.
    \item {\em Security}: at sensitive levels, data may be limited to a small subset of enterprise services and operations staff for network ingress, typically in ``break glass'' scenarios~\cite{brucker2009}.
\end{compactitem}

\descr{Data Access for Training.}
Model training necessitates data access, posing a significant security challenge to address. In addition to adhering to best industry best practices including {\em principle of least privilege}~\cite{chen2007} and {\em defense in depth}, data access poses further challenges:
\begin{compactitem}
    \item {\em Training at source}: training executed at or nearby the data within the same security tier.
    \item {\em Temporary privilege escalation}: short-lived delegation of control to ephemeral compute tasks.
    \item {\em Avoid data leakage}: ensuring all computation is done in-memory without spilling to disk.
    \item {\em Encryption}: all intermediate data encrypted at-rest.
\end{compactitem}

\descr{Data/Model Security.}
There is an asymmetry in security provisions between data at-rest environments and compute environments used for synthetic data model training and generation.
This presents an architectural challenge, and solutions may vary across different security contexts.

\descr{Data/Model Management.}
Synthetic data integration into enterprise data systems causes several challenges:
\begin{compactitem}
    \item {\em Versioning}: data versioning, synchronization, refreshment and metadata management~\cite{nargesian2019data} are all important for up-to-date synthetic data.
    \item {\em Compatibility}: model portability across environments and synthetic data compatibility with various locations/formats both need to be considered.
    \item {\em Compute resources}: high resource requirements for model training and data storage necessitate load distribution across multiple machines and efficient management of compute-intensive resource allocations.
\end{compactitem}

\subsection{Governance}
\label{sec:gov}
As enterprises deploy synthetic data, careful governance of the lifeycle processes and efficiency including analysis, policies, communication, and scaling become crucial.

\descr{AI Governance.}
AI governance is still in early stages, which poses challenges for enterprises using synthetic data models to develop governance frameworks:
\begin{compactitem}
    \item {\em Cross-team collaboration}:
    creating a unified governance framework is challenging when involving multiple stakeholders, including i) model builders (data scientists/engineers), ii) data owners/controllers, domain experts, and iii) governing stakeholders (legal/privacy experts)~\cite{ai_gov_global, pwc_risk_gen_ai}.
    \item {\em Limited monitoring}: the lack of industry standards for monitoring synthetic data models makes tool building and/or selection difficult.
\end{compactitem}

\descr{Scalability.}
Enterprise segmentation and diverse legislation across countries present challenges due to limited knowledge exchange and differing regulatory requirements:
\begin{compactitem}
    \item {\em Access control}: complex team structures hinder i) tracking ownership of data used by synthetic data models, ii) implementing precise access control, and iii) preventing data exploitation (e.g., proper disposal of data and the copy problem~\cite{trask2020beyond}).
    \item {\em Security}: models should be tested for potential vulnerabilities such as data poisoning, query injection attacks.
    \item {\em Policies}: data governance frameworks, privacy policies, and compliance procedures are resource-intensive to create and disseminate.
    \item {\em Environment controls}: clear internal guidance on the operational scope of synthetic data models is crucial for enterprises with multiple environments (e.g., staging vs production) of varying data sensitivity levels.
\end{compactitem}

\descr{Ethics.}
Aligning on the ethics of synthetic data at various organizational levels can be challenging and time-consuming, considering privacy, explainability, fairness, agency, and sourcing considerations at scale ~\cite{uksa2022ethical}.
\begin{compactitem}
    \item {\em Explainability and interpretability}: ensuring synthetic data explainability is challenging due to ongoing internal governance, knowledge, and research gaps, yet it plays a crucial role in instilling trust~\cite{ohm2009broken}.
    \item {\em Transparency}: obtaining explicit user consent and publicizing synthetic data policies builds trust but opens up risk of scrutiny and reputational damage.
\end{compactitem}

\subsection{Compliance \& Regulation}
Large corporations are facing growing difficulties in adhering to regulations governing data protection and AI laws.

\descr{Landscape.}
On the one hand, over 120 countries have adopted data protection and AI legislation~\cite{edps2022data} (e.g., GDPR, HIPAA, AI Act) while regulators are issuing companies more/higher-value fines~\cite{cms2023gdpr}.
On the other, enterprises are caught between laws balancing innovation, competition, ethical considerations, and regulatory compliance.
\citet{gal2023synthetic} warn that synthetic data could exacerbate these challenges.

\descr{Data Protection.}
Understanding and adhering to regulation is time/resource intensive.
Main considerations include:
\begin{compactitem}
    \item {\em Privacy engineering}: integrating privacy protections across the design and development cycles include privacy by design, data minimization, informed consent, accountability, transparency, purpose limitation.
    \item {\em Anonymization}: \citet{eu2014opinion, ico2021how} assert that {\em singling out}, {\em linkability}, and {\em inferences} risks need to be reduced for sufficient anonymization under the lenses of {\em motivated intruder} test.
    Yet there is not enough clarity or concrete guidelines~\cite{fca2023synthetic}.
\end{compactitem}

\descr{Tech-Legal Gap.}
Mapping technology to legal concepts, and vice versa, is a complex process requiring careful consideration and expertise.
\citet{bellovin2019privacy, lopez2022on, ganev2023synthetic} argue that applying DP to synthetic data techniques can address privacy concerns, confirmed by empirically evaluations~\cite{giomi2022unified}.

\subsection{Adoption}
\label{sec:adop}

In recent years, the number of AI capabilities used by organizations has more than doubled~\cite{mckinsey2022ai}.
In this section, we highlight the key challenges when driving synthetic data adoption at scale.

\descr{Strategic Alignment \& Outcomes.}
Unspecified metrics and ROI measures, along with a lack of strategic alignment, can hinder achieving business outcomes:
\begin{compactitem}
    \item {\em Vision \& strategy}:
    without a clear synthetic data vision, aligned with the broader business strategy, behaviors and target outcomes are unlikely to materialize.
    Alignment across the business on objectives will drive more measurable value~\cite{accenture2021}.
    \item {\em Business metrics \& ROI}:
    misalignment on business metrics and ROI measures can jeopardize decision-making, investment, and performance tracking.
\end{compactitem}

\descr{Operational Effectiveness.}
Operational challenges include workflow adaptation, team training, and cost management:
\begin{compactitem}
    \item {\em Existing workflows}:
    New technologies will create fundamental changes in workflows, roles, and culture~\cite{HBR2019}.
    Optimizing producer/consumer workflows is crucial for efficient synthetic data generation, utilization, and feedback loops.
    \item {\em Capability ownership}: a lack of coordination managed by a central team can slow adoption due to conflicting priorities and communication issues.
    \item {\em Skills gap \& training}: Organizations are struggling to find employees with the combination of skills and knowledge to unleash the full potential of AI~\cite{MITProfedu}.
    Thus, adequate upskilling of synthetic data users is essential to address knowledge disparities and aid adoption.
    \item {\em Cost management}:
    implementing synthetic data at scale incurs costs (e.g., monetary investments, resource allocation, infrastructure requirements) that must be carefully managed before benefits are realized.
\end{compactitem}

\descr{Organizational Change.}
User skepticism, stakeholder buy-in, and effective communication impact change:
\begin{compactitem}
    \item {\em User scepticism}: growing scrutiny and concerns over Generative AI could hinder synthetic data adoption.
    \item {\em Change management}: successful transformation programs shift from a sole focus on technology to account for the human experience~\cite{EY2021}. As synthetic data becomes integrated across the enterprise, failure to guide organizational change and adapt the culture can impede adoption.

    \item {\em Leadership \& influencer buy-in}: overlooking leadership buy-in and cultural influencers can reduce confidence and trust in a change program~\cite{hbr2023getting}.
\end{compactitem}

\section{Discussion}
\label{sec:dis}

Given the wide-ranging applications and benefits of synthetic data, there are technical, architectural, governance, and adoption challenges related to deploying it within an enterprise.
However, these challenges are not obstructions, nor do they need to be addressed simultaneously.

\descr{State of Play.}
Unlike nascent Generative AI technologies, synthetic data is delivering on its initial promises in commercial environments.
The focus now turns to how it can be adopted at scale as a core technology in an enterprise data strategy.
Early adopters have invested in proofs of concept (POCs) that have demonstrated tangible business value, such as increased efficiency, accelerated innovation, and reduced compliance risk~\cite{benedetto2018creation, nature2023synthetic} but deployment and adoption should be approached holistically and prioritised based on strategic objectives.

\descr{Structured Approach.}
Organizations can adopt a structured approach to anchor their synthetic data programs, which includes assessing external dependencies, guiding activities related to deployment, governance, and adoption.
By involving diverse stakeholders and the appropriate expertise, focusing initially on low-risk use cases with quick time to value, and creating awareness about synthetic data, organizations can gain buy-in and establish trust.

Finally, in App.~\ref{sec:dep}, we outline a simplified three-staged process which prioritizes focus areas to address.

\section{Conclusion}

We have identified and categorized numerous challenges associated with large-scale deployment of synthetic data.
We believe our work will be valuable for practitioners and professionals interested in adopting synthetic data solutions.

\newpage
{\small
%\bibliography{mybib}

\begin{thebibliography}{77}
\providecommand{\natexlab}[1]{#1}
\providecommand{\url}[1]{\texttt{#1}}
\expandafter\ifx\csname urlstyle\endcsname\relax
  \providecommand{\doi}[1]{doi: #1}\else
  \providecommand{\doi}{doi: \begingroup \urlstyle{rm}\Url}\fi

\bibitem[{A29WP}(2014)]{eu2014opinion}
{A29WP}.
\newblock {Opinion on anonymisation techniques}.
\newblock
  \url{https://ec.europa.eu/justice/article-29/documentation/opinion-recommendation/files/2014/wp216_en.pdf},
  2014.

\bibitem[{Accenture}(2021)]{accenture2021}
{Accenture}.
\newblock {Aligning data strategy and digital transformation}.
\newblock
  \url{https://www.accenture.com/hk-en/insights/strategy/aligning-data-strategy-digital-transformation},
  2021.

\bibitem[Alaa et~al.(2022)Alaa, Van~Breugel, Saveliev, and van~der
  Schaar]{alaa2022faithful}
Alaa, A., Van~Breugel, B., Saveliev, E.~S., and van~der Schaar, M.
\newblock {How faithful is your synthetic data? sample-level metrics for
  evaluating and auditing generative models}.
\newblock In \emph{ICML}, 2022.

\bibitem[Assefa et~al.(2020)Assefa, Dervovic, Mahfouz, Tillman, Reddy, and
  Veloso]{assefa2020generating}
Assefa, S.~A., Dervovic, D., Mahfouz, M., Tillman, R.~E., Reddy, P., and
  Veloso, M.
\newblock {Generating synthetic data in finance: opportunities, challenges and
  pitfalls}.
\newblock In \emph{ACM ICAIF}, 2020.

\bibitem[Avram(2014)]{avram2014}
Avram, M.~G.
\newblock {Advantages and challenges of adopting cloud computing from an
  enterprise perspective}.
\newblock \emph{Procedia Technology}, 2014.

\bibitem[Bano et~al.(2016)Bano, Zowghi, and Sarkissian]{bano2016empirical}
Bano, M., Zowghi, D., and Sarkissian, N.
\newblock {Empirical study of communication structures and barriers in
  geographically distributed teams}.
\newblock \emph{IET software}, 2016.

\bibitem[Bellovin et~al.(2019)Bellovin, Dutta, and
  Reitinger]{bellovin2019privacy}
Bellovin, S.~M., Dutta, P.~K., and Reitinger, N.
\newblock {Privacy and synthetic datasets}.
\newblock \emph{STLR}, 2019.

\bibitem[Benedetto et~al.(2018)Benedetto, Stanley, Totty,
  et~al.]{benedetto2018creation}
Benedetto, G., Stanley, J.~C., Totty, E., et~al.
\newblock {The creation and use of the SIPP synthetic Beta v7. 0}.
\newblock \emph{US Census Bureau}, 2018.

\bibitem[Brucker \& Petritsch(2009)Brucker and Petritsch]{brucker2009}
Brucker, A.~D. and Petritsch, H.
\newblock {Extending access control models with break-glass}.
\newblock In \emph{ACM SACMAT}, 2009.

\bibitem[Butcher \& Beridze(2019)Butcher and Beridze]{ai_gov_global}
Butcher, J. and Beridze, I.
\newblock {What is the State of Artificial Intelligence Governance Globally?}
\newblock \emph{The RUSI Journal}, 2019.

\bibitem[Carlini et~al.(2019)Carlini, Liu, Erlingsson, Kos, and
  Song]{carlini2019secret}
Carlini, N., Liu, C., Erlingsson, {\'U}., Kos, J., and Song, D.
\newblock {The secret sharer: Evaluating and testing unintended memorization in
  neural networks}.
\newblock In \emph{USENIX Security}, 2019.

\bibitem[Carlini et~al.(2021)Carlini, Tramer, Wallace, Jagielski, Herbert-Voss,
  Lee, Roberts, Brown, Song, Erlingsson, Oprea, and
  Raffel]{carlini2021extracting}
Carlini, N., Tramer, F., Wallace, E., Jagielski, M., Herbert-Voss, A., Lee, K.,
  Roberts, A., Brown, T., Song, D., Erlingsson, U., Oprea, A., and Raffel, C.
\newblock {Extracting training data from large language models}.
\newblock In \emph{USENIX Security}, 2021.

\bibitem[Chen \& Crampton(2007)Chen and Crampton]{chen2007}
Chen, L. and Crampton, J.
\newblock {Inter-domain role mapping and least privilege}.
\newblock In \emph{ACM SACMAT}, 2007.

\bibitem[{CMS}(2023)]{cms2023gdpr}
{CMS}.
\newblock {GDPR Enforcement Tracker}.
\newblock \url{https://www.enforcementtracker.com/}, 2023.

\bibitem[{CNN}(2023)]{cnn2023dont}
{CNN}.
\newblock {Don’t tell anything to a chatbot you want to keep private}.
\newblock
  \url{https://edition.cnn.com/2023/04/06/tech/chatgpt-ai-privacy-concerns/index.html},
  2023.

\bibitem[Cummings et~al.(2023)Cummings, Desfontaines, Evans, Geambasu,
  Jagielski, Huang, Kairouz, Kamath, Oh, Ohrimenko, Papernot, Rogers, Shen,
  Song, Su, Terzis, Thakurta, Vassilvitskii, Wang, Xiong, Yekhanin, Yu, Zhan,
  and Zhang]{cummings2023challenges}
Cummings, R., Desfontaines, D., Evans, D., Geambasu, R., Jagielski, M., Huang,
  Y., Kairouz, P., Kamath, G., Oh, S., Ohrimenko, O., Papernot, N., Rogers, R.,
  Shen, M., Song, S., Su, W., Terzis, A., Thakurta, A., Vassilvitskii, S.,
  Wang, Y.-X., Xiong, L., Yekhanin, S., Yu, D., Zhan, H., and Zhang, W.
\newblock {Challenges towards the Next Frontier in Privacy}.
\newblock \emph{arXiv:2304.06929}, 2023.

\bibitem[Dwork \& Roth(2014)Dwork and Roth]{dwork2014algorithmic}
Dwork, C. and Roth, A.
\newblock {The algorithmic foundations of differential privacy}.
\newblock \emph{Foundations and Trends in Theoretical Computer Science}, 2014.

\bibitem[{EDPS}(2022)]{edps2022data}
{EDPS}.
\newblock {Data Protection concerns all of us}.
\newblock
  \url{https://edps.europa.eu/press-publications/press-news/blog/data-protection-concerns-all-us_en},
  2022.

\bibitem[{Entrepreneur}(2023)]{entrepreneur2023history}
{Entrepreneur}.
\newblock {History Has Shown What Happens to Companies that Shy Away from New
  Tech, So Why Are So Many Afraid of Generative AI?}
\newblock
  \url{https://www.entrepreneur.com/leadership/why-are-so-many-companies-afraid-of-generative-ai/446198},
  2023.

\bibitem[{EP and Council}(2016)]{official2016article}
{EP and Council}.
\newblock {Article 4 GDPR Definitions}.
\newblock \url{https://gdpr-info.eu/art-4-gdpr/}, 2016.

\bibitem[{EY}(2021)]{EY2021}
{EY}.
\newblock {Why change management is crucial during technologic
  transformations}.
\newblock
  \url{https://www.ey.com/en_be/consulting/why-change-management-is-crucial-during-technologic-transformations},
  2021.

\bibitem[{FCA}(2023)]{fca2023synthetic}
{FCA}.
\newblock {Synthetic data call for input feedback statement}.
\newblock \url{https://www.fca.org.uk/publication/feedback/fs23-1.pdf}, 2023.

\bibitem[Fowler(2012)]{fowler2012}
Fowler, M.
\newblock \emph{{Patterns of Enterprise Application Architecture}}.
\newblock Pearson Education, 2012.

\bibitem[Gal \& Lynskey(2023)Gal and Lynskey]{gal2023synthetic}
Gal, M. and Lynskey, O.
\newblock {Synthetic Data: Legal Implications of the Data-Generation
  Revolution}.
\newblock \emph{109 Iowa Law Review}, 2023.

\bibitem[Ganev(2023)]{ganev2023synthetic}
Ganev, G.
\newblock {When synthetic data met regulation}.
\newblock In \emph{ICML Workshop on Generative AI and Law}, 2023.

\bibitem[Ganev et~al.(2022)Ganev, Oprisanu, and De~Cristofaro]{ganev2022robin}
Ganev, G., Oprisanu, B., and De~Cristofaro, E.
\newblock {Robin Hood and Matthew Effects: Differential privacy has disparate
  impact on synthetic data}.
\newblock In \emph{ICML}, 2022.

\bibitem[Ganev et~al.(2023)Ganev, Xu, and
  De~Cristofaro]{ganev2023understanding}
Ganev, G., Xu, K., and De~Cristofaro, E.
\newblock {Understanding how Differentially Private Generative Models Spend
  their Privacy Budget}.
\newblock \emph{arXiv:2305.10994}, 2023.

\bibitem[{Gartner}(2021)]{gartner2021data}
{Gartner}.
\newblock {Data Sharing Is a Business Necessity to Accelerate Digital
  Business}.
\newblock
  \url{https://www.gartner.com/smarterwithgartner/data-sharing-is-a-business-necessity-to-accelerate-digital-business},
  2021.

\bibitem[{Gartner}(2023)]{gartner2023beyond}
{Gartner}.
\newblock {Beyond ChatGPT: The Future of Generative AI for Enterprises}.
\newblock
  \url{https://www.gartner.com/en/articles/beyond-chatgpt-the-future-of-generative-ai-for-enterprises},
  2023.

\bibitem[Giomi et~al.(2022)Giomi, Boenisch, Wehmeyer, and
  Tasn{\'a}di]{giomi2022unified}
Giomi, M., Boenisch, F., Wehmeyer, C., and Tasn{\'a}di, B.
\newblock {A unified framework for quantifying privacy risk in synthetic data}.
\newblock In \emph{PETs}, 2022.

\bibitem[Goodfellow et~al.(2014)Goodfellow, Pouget-Abadie, Mirza, Xu,
  Warde-Farley, Ozair, Courville, and Bengio]{goodfellow2014generative}
Goodfellow, I., Pouget-Abadie, J., Mirza, M., Xu, B., Warde-Farley, D., Ozair,
  S., Courville, A., and Bengio, Y.
\newblock {Generative adversarial nets}.
\newblock \emph{NIPS}, 2014.

\bibitem[Hayes et~al.(2019)Hayes, Melis, Danezis, and
  De~Cristofaro]{hayes2019logan}
Hayes, J., Melis, L., Danezis, G., and De~Cristofaro, E.
\newblock {Logan: membership inference attacks against generative models}.
\newblock In \emph{PoPETs}, 2019.

\bibitem[{HBR}(2019)]{HBR2019}
{HBR}.
\newblock {Building the AI-Powered Organization}.
\newblock \url{https://hbr.org/2019/07/building-the-ai-powered-organization},
  2019.

\bibitem[{HBR}(2023)]{hbr2023getting}
{HBR}.
\newblock {Getting Employee Buy-In for Organizational Change}.
\newblock
  \url{https://hbr.org/2023/02/getting-employee-buy-in-for-organizational-change},
  2023.

\bibitem[Hilprecht et~al.(2019)Hilprecht, H{\"a}rterich, and
  Bernau]{hilprecht2019monte}
Hilprecht, B., H{\"a}rterich, M., and Bernau, D.
\newblock {Monte Carlo and Reconstruction Membership Inference Attacks against
  Generative Models}.
\newblock In \emph{PoPETs}, 2019.

\bibitem[Houssiau et~al.(2022{\natexlab{a}})Houssiau, Cohen, Szpruch, Daniel,
  Lawrence, Mitra, Wilde, and Mole]{houssiau2022framework}
Houssiau, F., Cohen, S.~N., Szpruch, L., Daniel, O., Lawrence, M.~G., Mitra,
  R., Wilde, H., and Mole, C.
\newblock {A Framework for Auditable Synthetic Data Generation}.
\newblock \emph{arXiv:2211.11540}, 2022{\natexlab{a}}.

\bibitem[Houssiau et~al.(2022{\natexlab{b}})Houssiau, Jordon, Cohen, Daniel,
  Elliott, Geddes, Mole, Rangel-Smith, and Szpruch]{houssiau2022tapas}
Houssiau, F., Jordon, J., Cohen, S.~N., Daniel, O., Elliott, A., Geddes, J.,
  Mole, C., Rangel-Smith, C., and Szpruch, L.
\newblock {TAPAS: a toolbox for adversarial privacy auditing of synthetic
  data}.
\newblock In \emph{NeurIPS Workshop on SyntheticData4ML}, 2022{\natexlab{b}}.

\bibitem[Hsu et~al.(2014)Hsu, Gaboardi, Haeberlen, Khanna, Narayan, Pierce, and
  Roth]{hsu2014differential}
Hsu, J., Gaboardi, M., Haeberlen, A., Khanna, S., Narayan, A., Pierce, B.~C.,
  and Roth, A.
\newblock {Differential privacy: an economic method for choosing epsilon}.
\newblock In \emph{IEEE CSF}, 2014.

\bibitem[{ICO}(2022)]{ico2022privacy}
{ICO}.
\newblock {Chapter 5: privacy-enhancing technologies (PETs)}.
\newblock
  \url{https://ico.org.uk/media/about-the-ico/consultations/4021464/chapter-5-anonymisation-pets.pdf},
  2022.

\bibitem[{ICO UK}(2021)]{ico2021how}
{ICO UK}.
\newblock {Chapter 2: how do we ensure anonymisation is effective?}
\newblock
  \url{https://ico.org.uk/media/about-the-ico/documents/4018606/chapter-2-anonymisation-draft.pdf},
  2021.

\bibitem[Jamshidi et~al.(2013)Jamshidi, Ahmad, and Pahl]{jamshidi2013}
Jamshidi, P., Ahmad, A., and Pahl, C.
\newblock {Cloud migration research: a systematic review}.
\newblock \emph{IEEE TCC}, 2013.

\bibitem[Jordon et~al.(2018)Jordon, Yoon, and Van Der~Schaar]{jordon2018pate}
Jordon, J., Yoon, J., and Van Der~Schaar, M.
\newblock {PATE-GAN: Generating synthetic data with differential privacy
  guarantees}.
\newblock In \emph{ICLR}, 2018.

\bibitem[Jordon et~al.(2022)Jordon, Szpruch, Houssiau, Bottarelli, Cherubin,
  Maple, Cohen, and Weller]{jordon2022synthetic}
Jordon, J., Szpruch, L., Houssiau, F., Bottarelli, M., Cherubin, G., Maple, C.,
  Cohen, S.~N., and Weller, A.
\newblock {Synthetic Data--what, why and how?}
\newblock \emph{arXiv:2205.03257}, 2022.

\bibitem[{KPMG}(2023)]{kpmg2023kpmg}
{KPMG}.
\newblock {KPMG Generative AI Survey}.
\newblock
  \url{https://info.kpmg.us/news-perspectives/technology-innovation/kpmg-generative-ai-2023.html},
  2023.

\bibitem[Leffingwell(2007)]{leffingwell2007scaling}
Leffingwell, D.
\newblock \emph{{Scaling software agility: best practices for large
  enterprises}}.
\newblock Pearson Education, 2007.

\bibitem[Lin et~al.(2020)Lin, Jain, Wang, Fanti, and Sekar]{lin2020using}
Lin, Z., Jain, A., Wang, C., Fanti, G., and Sekar, V.
\newblock {Using gans for sharing networked time series data: Challenges,
  initial promise, and open questions}.
\newblock In \emph{ACM IMC}, 2020.

\bibitem[L{\'o}pez \& Elbi(2022)L{\'o}pez and Elbi]{lopez2022on}
L{\'o}pez, C. A.~F. and Elbi, A.
\newblock {On the legal nature of synthetic data}.
\newblock In \emph{NeurIPS SyntheticData4ML}, 2022.

\bibitem[McKenna et~al.(2021)McKenna, Miklau, and Sheldon]{mckenna2021winning}
McKenna, R., Miklau, G., and Sheldon, D.
\newblock {Winning the NIST Contest: A scalable and general approach to
  differentially private synthetic data}.
\newblock \emph{JPC}, 2021.

\bibitem[McKenna et~al.(2022)McKenna, Mullins, Sheldon, and
  Miklau]{mckenna2022aim}
McKenna, R., Mullins, B., Sheldon, D., and Miklau, G.
\newblock {Aim: An adaptive and iterative mechanism for differentially private
  synthetic data}.
\newblock \emph{PVLDB}, 2022.

\bibitem[{McKinsey}(2019)]{mckinsey2019driving}
{McKinsey}.
\newblock {Driving impact at scale from automation and AI}.
\newblock
  \url{https://www.mckinsey.com/~/media/McKinsey/Business\%20Functions/McKinsey\%20Digital/Our\%20Insights/Driving\%20impact\%20at\%20scale\%20from\%20automation\%20and\%20AI/Driving-impact-at-scale-from-automation-and-AI.ashx},
  2019.

\bibitem[{McKinsey}(2022)]{mckinsey2022ai}
{McKinsey}.
\newblock {The state of AI in 2022—and a half decade in review}.
\newblock
  \url{https://www.mckinsey.com/capabilities/quantumblack/our-insights/the-state-of-ai-in-2022-and-a-half-decade-in-review#review},
  2022.

\bibitem[{MIT Professional}(2018)]{MITProfedu}
{MIT Professional}.
\newblock {Four key barriers to the widespread adoption of AI}.
\newblock
  \url{https://professional.mit.edu/news/articles/four-key-barriers-widespread-adoption-ai},
  2018.

\bibitem[Nargesian et~al.(2019)Nargesian, Zhu, Miller, Pu, and
  Arocena]{nargesian2019data}
Nargesian, F., Zhu, E., Miller, R.~J., Pu, K.~Q., and Arocena, P.~C.
\newblock {Data Lake Management: Challenges and Opportunities}.
\newblock \emph{PVLDB}, 2019.

\bibitem[{Nature}(2023)]{nature2023synthetic}
{Nature}.
\newblock {Synthetic data could be better than real data}.
\newblock \url{https://www.nature.com/articles/d41586-023-01445-8}, 2023.

\bibitem[{NYT}(2018)]{nyt2018as}
{NYT}.
\newblock {As Facebook Raised a Privacy Wall, It Carved an Opening for Tech
  Giants}.
\newblock
  \url{https://www.nytimes.com/2018/12/18/technology/facebook-privacy.html},
  2018.

\bibitem[{NYT}(2019)]{nyt2019twelve}
{NYT}.
\newblock {Twelve Million Phones, One Dataset, Zero Privacy}.
\newblock
  \url{https://www.nytimes.com/interactive/2019/12/19/opinion/location-tracking-cell-phone.html},
  2019.

\bibitem[OECD(2023)]{oecd2023emerging}
OECD.
\newblock Emerging privacy-enhancing technologies.
\newblock \url{https://www.oecd-ilibrary.org/content/paper/bf121be4-en}, 2023.

\bibitem[Ohm(2009)]{ohm2009broken}
Ohm, P.
\newblock {Broken promises of privacy: Responding to the surprising failure of
  anonymization}.
\newblock \emph{UCLA Law Review}, 2009.

\bibitem[{PCI SSC}(2013)]{pci2013}
{PCI SSC}.
\newblock {Information Supplement: PCI DSS Cloud Computing Guidelines}.
\newblock
  \url{https://listings.pcisecuritystandards.org/pdfs/PCI_DSS_v2_Cloud_Guidelines.pdf},
  2013.

\bibitem[{PwC}(2023)]{pwc_risk_gen_ai}
{PwC}.
\newblock {Managing the risk of generative AI}.
\newblock
  \url{https://explore.pwc.com/generativeai?_pfses=D8nsC9bP5NQMW25zxpYx69tC},
  2023.

\bibitem[{Reuters}(2023)]{reuters2023chatgpt}
{Reuters}.
\newblock {ChatGPT sets record for fastest-growing user base}.
\newblock
  \url{https://www.reuters.com/technology/chatgpt-sets-record-fastest-growing-user-base-analyst-note-2023-02-01/},
  2023.

\bibitem[{Royal Society}(2023)]{rs2023privacy}
{Royal Society}.
\newblock {From privacy to partnership: the role of PETs in data governance and
  collaborative analysis}.
\newblock
  \url{https://royalsociety.org/-/media/policy/projects/privacy-enhancing-technologies/From-Privacy-to-Partnership.pdf},
  2023.

\bibitem[{Sequoia Capital}(2022)]{sequoia2022generative}
{Sequoia Capital}.
\newblock {Generative AI: A Creative New World}.
\newblock
  \url{https://www.sequoiacap.com/article/generative-ai-a-creative-new-world/},
  2022.

\bibitem[Sohl-Dickstein et~al.(2015)Sohl-Dickstein, Weiss, Maheswaranathan, and
  Ganguli]{sohl2015deep}
Sohl-Dickstein, J., Weiss, E., Maheswaranathan, N., and Ganguli, S.
\newblock {Deep unsupervised learning using nonequilibrium thermodynamics}.
\newblock In \emph{ICML}, 2015.

\bibitem[Stadler et~al.(2022)Stadler, Oprisanu, and
  Troncoso]{stadler2022synthetic}
Stadler, T., Oprisanu, B., and Troncoso, C.
\newblock {Synthetic data -- anonymization groundhog day}.
\newblock In \emph{Usenix Security}, 2022.

\bibitem[Tao et~al.(2022)Tao, McKenna, Hay, Machanavajjhala, and
  Miklau]{tao2022benchmarking}
Tao, Y., McKenna, R., Hay, M., Machanavajjhala, A., and Miklau, G.
\newblock {Benchmarking differentially private synthetic data generation
  algorithms}.
\newblock \emph{PPAI}, 2022.

\bibitem[{TechCrunch}(2023{\natexlab{a}})]{techcrunch2023apple}
{TechCrunch}.
\newblock {Apple reportedly limits internal use of AI-powered tools like
  ChatGPT and GitHub Copilot}.
\newblock
  \url{https://techcrunch.com/2023/05/19/apple-reportedly-limits-internal-use-of-ai-powered-tools-like-chatgpt-and-github-copilot/},
  2023{\natexlab{a}}.

\bibitem[{TechCrunch}(2023{\natexlab{b}})]{techcrunch2023the}
{TechCrunch}.
\newblock {The current legal cases against generative AI are just the
  beginning}.
\newblock
  \url{https://techcrunch.com/2023/01/27/the-current-legal-cases-against-generative-ai-are-just-the-beginning/},
  2023{\natexlab{b}}.

\bibitem[Trask et~al.(2020)Trask, Bluemke, Garfinkel, Cuervas-Mons, and
  Dafoe]{trask2020beyond}
Trask, A., Bluemke, E., Garfinkel, B., Cuervas-Mons, C.~G., and Dafoe, A.
\newblock {Beyond privacy trade-offs with structured transparency}.
\newblock \emph{arXiv:2012.08347}, 2020.

\bibitem[{UKSA}(2022)]{uksa2022ethical}
{UKSA}.
\newblock {Ethical considerations relating to the creation and use of synthetic
  data}.
\newblock
  \url{https://uksa.statisticsauthority.gov.uk/publication/ethical-considerations-relating-to-the-creation-and-use-of-synthetic-data/},
  2022.

\bibitem[{UN}(2023)]{un2023guide}
{UN}.
\newblock {The United Nations Guide on privacy-enhancing technologies for
  official statistics}.
\newblock
  \url{https://unstats.un.org/bigdata/task-teams/privacy/guide/2023_UN\%20PET\%20Guide.pdf},
  2023.

\bibitem[van Breugel \& van~der Schaar(2023)van Breugel and van~der
  Schaar]{breugel2023beyond}
van Breugel, B. and van~der Schaar, M.
\newblock {Beyond Privacy: Navigating the Opportunities and Challenges of
  Synthetic Data}.
\newblock \emph{arXiv:2304.03722}, 2023.

\bibitem[Vaswani et~al.(2017)Vaswani, Shazeer, Parmar, Uszkoreit, Jones, Gomez,
  Kaiser, and Polosukhin]{vaswani2017attention}
Vaswani, A., Shazeer, N., Parmar, N., Uszkoreit, J., Jones, L., Gomez, A.~N.,
  Kaiser, {\L}., and Polosukhin, I.
\newblock {Attention is all you need}.
\newblock \emph{NeurIPS}, 2017.

\bibitem[Weidinger et~al.(2021)Weidinger, Mellor, Rauh, Griffin, Uesato, Huang,
  Cheng, Glaese, Balle, Kasirzadeh, Kenton, Brown, Hawkins, Stepleton, Biles,
  Birhane, Haas, Rimell, Hendricks, Isaac, Legassick, Irving, and
  Gabriel]{weidinger2021ethical}
Weidinger, L., Mellor, J., Rauh, M., Griffin, C., Uesato, J., Huang, P.-S.,
  Cheng, M., Glaese, M., Balle, B., Kasirzadeh, A., Kenton, Z., Brown, S.,
  Hawkins, W., Stepleton, T., Biles, C., Birhane, A., Haas, J., Rimell, L.,
  Hendricks, L.~A., Isaac, W., Legassick, S., Irving, G., and Gabriel, I.
\newblock {Ethical and social risks of harm from language models}.
\newblock \emph{arXiv:2112.04359}, 2021.

\bibitem[Xie et~al.(2018)Xie, Lin, Wang, Wang, and Zhou]{xie2018differentially}
Xie, L., Lin, K., Wang, S., Wang, F., and Zhou, J.
\newblock {Differentially private generative adversarial network}.
\newblock \emph{arXiv:1802.06739}, 2018.

\bibitem[Xu et~al.(2023)Xu, Ganev, Joubert, Davison, Van~Acker, and
  Robinson]{xu2023synthetic}
Xu, K., Ganev, G., Joubert, E., Davison, R., Van~Acker, O., and Robinson, L.
\newblock {Synthetic data generation of many-to-many datasets via random graph
  generation}.
\newblock In \emph{ICLR}, 2023.

\bibitem[Zhang et~al.(2017)Zhang, Cormode, Procopiuc, Srivastava, and
  Xiao]{zhang2017privbayes}
Zhang, J., Cormode, G., Procopiuc, C.~M., Srivastava, D., and Xiao, X.
\newblock {Privbayes: Private data release via bayesian networks}.
\newblock \emph{ACM TODS}, 2017.

\end{thebibliography}

\bibliographystyle{icml2023}
}
\newpage

\appendix

\section{Preliminaries}
\label{sec:prelim}

We introduce the foundational concepts that we use throughout the paper.

\descr{Enterprise.}
For the purposes of this paper, large enterprises (typically with over \$1B revenue and 100,000 customers) are established organizations with diverse business units and extensive legacy infrastructure.
They possess vast and diverse data, which is often distributed and inconsistent.
While they face burdensome regulations and the risk of reputational damage, their cautious approach to innovation is in contrast to the agility of startups and SMEs.

\descr{Personal and Proprietary Data.}
\citet{official2016article} define personal data as ``any information relating to an identified or identifiable natural person'' while the latter as someone who can directly or indirectly be identified, by reference to an identifier such as name, id number, location, etc.
On the other hand, proprietary data refers to privileged/confidential information owned by an organization that provides a competitive advantage (e.g., trade secrets, commercial/financial information) and is not public.

\descr{(Privacy-Preserving) Synthetic Data.}
Synthetic data is data generated by a purpose-built mathematical model or algorithm (i.e., machine learning generative model) trained on real data, with the aim of solving a (set of) task(s)~\cite{jordon2022synthetic}.
Unfortunately, unless models are trained with explicit privacy-preserving mechanisms, they could memorize and leak the privacy of input records~\cite{hayes2019logan, carlini2019secret, stadler2022synthetic}.
The state-of-the-art method to protect against such risks is to train models while satisfying Differential Privacy (DP)~\cite{dwork2014algorithmic}, whereby noisy/random mechanisms provably minimize the exposure of all records.
The level of exposure is quantified by an input parameter $\epsilon$, also known as the privacy budget.

We focus on privacy-preserving synthetic data, generated by DP models, even though synthetic data use cases could go beyond privacy (e.g., data augmentation, fairness, simulation, etc.~\cite{jordon2022synthetic, breugel2023beyond}).
However, privacy remains a central focus in all applications, given the need to handle sensitive data, which poses potential risks of financial loss, reputational damage, and loss of customers trust if mishandled.

\section{Deployment Phases}
\label{sec:dep}

As discussed in Sec.~\ref{sec:chall}, there are several challenges when deploying high quality enterprise synthetic data.
In this section, we outline a simplified three-stage approach and highlight the core areas to consider.

\subsection{Initial Phase}
Organizations embarking on their synthetic data journey can gain valuable insights from case studies and expert conversations with synthetic data professionals, including technology vendors, regulatory bodies, and system integrators.
\begin{compactitem}
    \item {\em Proofs of concept}: select initial use cases that will deliver strong proof points and quick time to value to prove the technology.
    Ensure the results and business outcomes are communicated with the relevant stakeholders to build awareness and advocacy, including how the synthetic data performs compared to previously used techniques (e.g., anonymization/masking).
    \item {\em Education}: as synthetic data is considered an emerging technology, education and foundational knowledge sharing is necessary for technical and non-technical audiences.
    An additional aim should be to increase visibility across the enterprise to spark further use cases.
    \item {\em Governance and metrics}:
    agree a core set of metrics to compare and showcase the results of synthetic data.
    Begin to consider governance frameworks and performance monitoring required for transitioning into the scaling phase.
\end{compactitem}

\subsection{Scaling Phase}
Many synthetic data early adopters are now in the Scaling Phase where the focus is on expansion.
This stage can be challenging for enterprises to navigate: delivery capacity has to grow to meet demand at the same time as the organisational change program and governance framework evolves to enable further adoption.
\begin{compactitem}
    \item {\em Strategic alignment}: as with any large-scale AI deployment, aligning the long-term business value of synthetic data with the broader strategy is crucial for setting direction and driving the right behaviours.
    The earlier key stakeholders are involved in this alignment exercise, the greater the chances of success in the later stages.

   \item {\em Regulation \& data tracking}: as adoption grows, organizations need to keep abreast of regulatory changes and internal governance should be updated accordingly -- covering both synthetic data models and synthetic data usage.
   This should be incorporated as part of the governance framework developed in the initial phase.
   \item {\em Scaling architecture}: to counter the complexity of data ownership and network topologies as noted in Sec.~\ref{sec:inf}, data flows should be mapped as much as possible before scaling synthetic data.
   \item {\em Monitoring \& auditing}: monitoring the models with logging, dashboards, and audit trails provide greater visibility across the enterprise and may be necessary for compliance.
\end{compactitem}

\subsection{Future Phase}
\label{sec:fut}
The focus areas in the Future Phase will be largely driven by the progress of synthetic data technologies, the evolving regulatory landscape and rate of adoption within enterprises.
Synthetic data will replace the use of real data across business areas as the production and consumption process becomes more streamlined.
\begin{compactitem}
    \item {\em Seamless \& integrated use}: as synthetic data is adopted across the enterprise, it should be embedded in the culture as a core component of the data strategy.
    There may be scope for more seamless ways to attain and consume synthetic data, for example, on-demand data and models via a marketplace.
   \item {\em Synthetic data advisory}: organizations will have established strong governance frameworks to reach this stage.
   They may wish to utilize knowledge and experience to advise external bodies and contribute to industry best practices.
\end{compactitem}

\end{document}